\documentclass[twocolumn,letterpaper]{IEEEAerospaceCLS}  %

\usepackage[]{graphicx}    %
\usepackage[colorlinks=true, urlcolor=blue, linkcolor=black, citecolor=black]{hyperref} %
\newcommand{\ignore}[1]{}  %
\usepackage{caption}
\usepackage{subcaption}
\usepackage{float}
\usepackage{dblfloatfix}

\usepackage{booktabs}
\usepackage{multirow}
\usepackage{amsmath}

\usepackage{xcolor}
\usepackage{xparse}
\usepackage[normalem]{ulem} %

\usepackage[
  backend=biber,
  style=ieee,
  maxbibnames=6,
  minbibnames=1,
  doi=false,isbn=false,url=false,eprint=false,
  eprint=false,
  url=true
]{biblatex}
\usepackage{xurl}

\AtEveryBibitem{%
  \ifentrytype{online}
    {}
    {%
      \clearfield{url}%
      \clearfield{urldate}%
    }%
}
\AtEveryBibitem{%
  \clearfield{urldate}
  \clearfield{urlyear}%
  \clearfield{urlmonth}%
  \clearfield{urlday}%
  \clearfield{note}%
}
\makeatletter
\def\bstctlcite{\@ifnextchar[{\@bstctlcite}{\@bstctlcite[@auxout]}}
\def\@bstctlcite[#1]#2{\@bsphack
  \@for\@citeb:=#2\do{%
    \edef\@citeb{\expandafter\@firstofone\@citeb}%
    \if@filesw
      \immediate\write\csname #1\endcsname{\string\citation{\@citeb}}%
    \fi}%
  \@esphack}
\makeatother

\usepackage{eso-pic}

\begin{document}
\bstctlcite{BSTcontrol}

\newif\ifedits \newif\iftodos
\NewDocumentCommand{\annotations}{m}{\csname ann#1\endcsname}
\newcommand{\annfull}{\editstrue\todostrue}
\newcommand{\annedits}{\editstrue\todosfalse}
\newcommand{\anntodos}{\editsfalse\todostrue}
\newcommand{\annclean}{\editsfalse\todosfalse}
\annotations{full}

\NewDocumentCommand{\add}{m}{\ifedits\textcolor{blue}{\textbf{[}#1\textbf{]}}\else#1\fi}
\NewDocumentCommand{\del}{m}{\ifedits\textcolor{red}{\sout{#1}}\fi}
\NewDocumentCommand{\rev}{mm}{\ifedits\textcolor{red}{\sout{#1}}\textcolor{blue}{#2}\else#2\fi}

\newcommand{\citeneeded}{\iftodos\textcolor{red}{\textbf{[X]}}\fi}
\newcommand{\verifythis}{\iftodos\textcolor{red}{\textbf{[VERIFY]}}\fi}
\NewDocumentCommand{\todo}{g}{\iftodos\textcolor{red}{\textbf{[TODO\IfValueT{#1}{: #1}]}}\fi}
\NewDocumentCommand{\note}{g}{\iftodos\textcolor{red}{\textbf{[NOTE\IfValueT{#1}{: #1}]}}\fi}

\newif\ifjournal
\newif\ifarxiv

\NewDocumentCommand{\paperMode}{m}{\csname mode#1\endcsname}
\newcommand{\modejournal}{\journaltrue\arxivfalse}
\newcommand{\modeconference}{\journalfalse\arxivfalse}
\newcommand{\modearxiv}{\journaltrue\arxivtrue}
\paperMode{journal}

\definecolor{journalgray}{gray}{0.35}
\definecolor{conferenceorange}{RGB}{200,100,0}

\newif\ifannotationsvisible
\newcommand{\setannotationsvisible}{%
  \annotationsvisiblefalse
  \ifedits\annotationsvisibletrue\fi
  \iftodos\annotationsvisibletrue\fi
}

\newcommand{\journalcolor}{%
  \setannotationsvisible
  \ifannotationsvisible
    \color{journalgray}%
  \fi
}

\newcommand{\conferencecolor}{%
  \setannotationsvisible
  \ifannotationsvisible
    \color{conferenceorange}%
  \fi
}

\NewDocumentEnvironment{journal}{+b}
  {%
    \setannotationsvisible
    \ifannotationsvisible
      \begingroup
      \journalcolor
      \captionsetup{
        font={color=journalgray},
        labelfont={color=journalgray}
      }%
      \let\journalfigure\figure
      \let\journaltable\table
      \RenewDocumentCommand{\figure}{o}{%
        \IfNoValueTF{##1}{\journalfigure}{\journalfigure[##1]}%
        \journalcolor
      }%
      \RenewDocumentCommand{\table}{o}{%
        \IfNoValueTF{##1}{\journaltable}{\journaltable[##1]}%
        \journalcolor
      }%
      #1%
      \endgroup
    \else
      \ifjournal
        #1%
      \fi
    \fi
  }
  {}

\NewDocumentEnvironment{conference}{+b}
  {%
    \ifarxiv
    \else
      \setannotationsvisible
      \ifannotationsvisible
        \begingroup
        \conferencecolor
        \captionsetup{
          font={color=conferenceorange},
          labelfont={color=conferenceorange}
        }%
        \let\conferencefigure\figure
        \let\conferencetable\table
        \RenewDocumentCommand{\figure}{o}{%
          \IfNoValueTF{##1}{\conferencefigure}{\conferencefigure[##1]}%
          \conferencecolor
        }%
        \RenewDocumentCommand{\table}{o}{%
          \IfNoValueTF{##1}{\conferencetable}{\conferencetable[##1]}%
          \conferencecolor
        }%
        #1%
        \endgroup
      \else
        \ifjournal
        \else
          #1%
        \fi
      \fi
    \fi
  }
  {}

\NewDocumentEnvironment{arxiv}{+b}
  {%
    \ifarxiv
      #1%
    \fi
  }
  {}

\NewDocumentEnvironment{notArxiv}{+b}
  {%
    \ifarxiv
    \else
      #1%
    \fi
  }
  {}

\NewDocumentCommand{\printAnnotationLegend}{}{%
  \begin{quote}
    \large

    \add{changes}

    \textcolor{journalgray}{gray text}
    = journal-version-only material.

    \textcolor{conferenceorange}{orange text}
    = conference-version-only material.
  \end{quote}
}

\annotations{clean}
\paperMode{arxiv}

\title{End-to-End Vision-Language-Action Policies for Autonomous Lunar Rover Docking}

\author{%
Sitan Huang, Grace Gao\\
Stanford University\\
William F. Durand Building\\
496 Lomita Mall\\
Stanford, CA 94305\\
\{huans16, gracegao\}@stanford.edu
\begin{notArxiv}
\thanks{\footnotesize 979-8-3195-1048-8/27/$\$31.00$ \copyright2027 IEEE}
\end{notArxiv}
}

\maketitle

\thispagestyle{plain}
\pagestyle{plain}

\begin{abstract}Surface navigation on the Moon is subject to extreme illumination contrast and deployment-time perceptual disturbances.
While modular navigation pipelines have long supported planetary rover navigation, robustness is compromised by measurement errors that propagate through hand-engineered components.
End-to-end learned policies are an alternative approach to navigation, which maps observations directly to control actions without intermediate components.
Vision-language-action models, in particular, have driven significant breakthroughs in robotics and autonomous vehicle research.
By leveraging internet-scale vision-language pretraining, these models can be exploited for their commonsense reasoning and visual understanding to provide robust generalization with limited task-specific data.
However, their applicability and performance have yet to be characterized for lunar surface operations.
In this study, we design a vision-language-action policy for autonomous lunar rover docking and characterize its performance relative to a modular navigation pipeline.
We develop a scalable training framework that generates safety-aware expert demonstrations in a high-fidelity lunar simulation environment.
In closed-loop evaluations, our policy performs competitively relative to the modular baseline under perceptual disturbances unseen during training%
\begin{journal}
.
\end{journal}%
\begin{conference}
\end{conference}
\begin{journal}
To address the tradeoff between compute consumption and closed-loop safety,
the policy adapts its execution horizon by determining how far the rover may safely travel in open loop.
Adaptive-horizon execution reduces inference frequency while limiting hazard-contact rates relative to fixed horizons in the evaluated scenarios.
\end{journal}

\end{abstract}

\tableofcontents

\section{Introduction}

Following the crewed Artemis II lunar flyby in 2026, sustained human exploration on the lunar surface is expected in the upcoming decade.
Large-scale surface missions will require substantial autonomy to limit operational costs and shorten mission cycles, using vehicles that autonomously offload, stage, relocate cargo, and dock with surface assets prior to crew arrival~\cite{nasa_lunar_2024,nasa_moon_2026}.

The lunar environment presents unique challenges for navigation.
Communication delays
and signal occlusion due to terrain make continuous ground control poorly suited for this task.
Lunar rovers must also operate in unmapped terrain with limited localization infrastructure. While proposed systems such as ESA’s Moonlight constellation will improve absolute positioning, local-relative navigation is necessary for hazard avoidance and safely maneuvering near astronauts and other vehicles.
Harsh lighting contrasts, long shadows, and feature-sparse terrain further complicate perception and navigation.

\begin{figure*}[t!]
    \centering
    \begin{subfigure}[h!]{0.50\textwidth}
        \centering
        \includegraphics[height=2.1in]{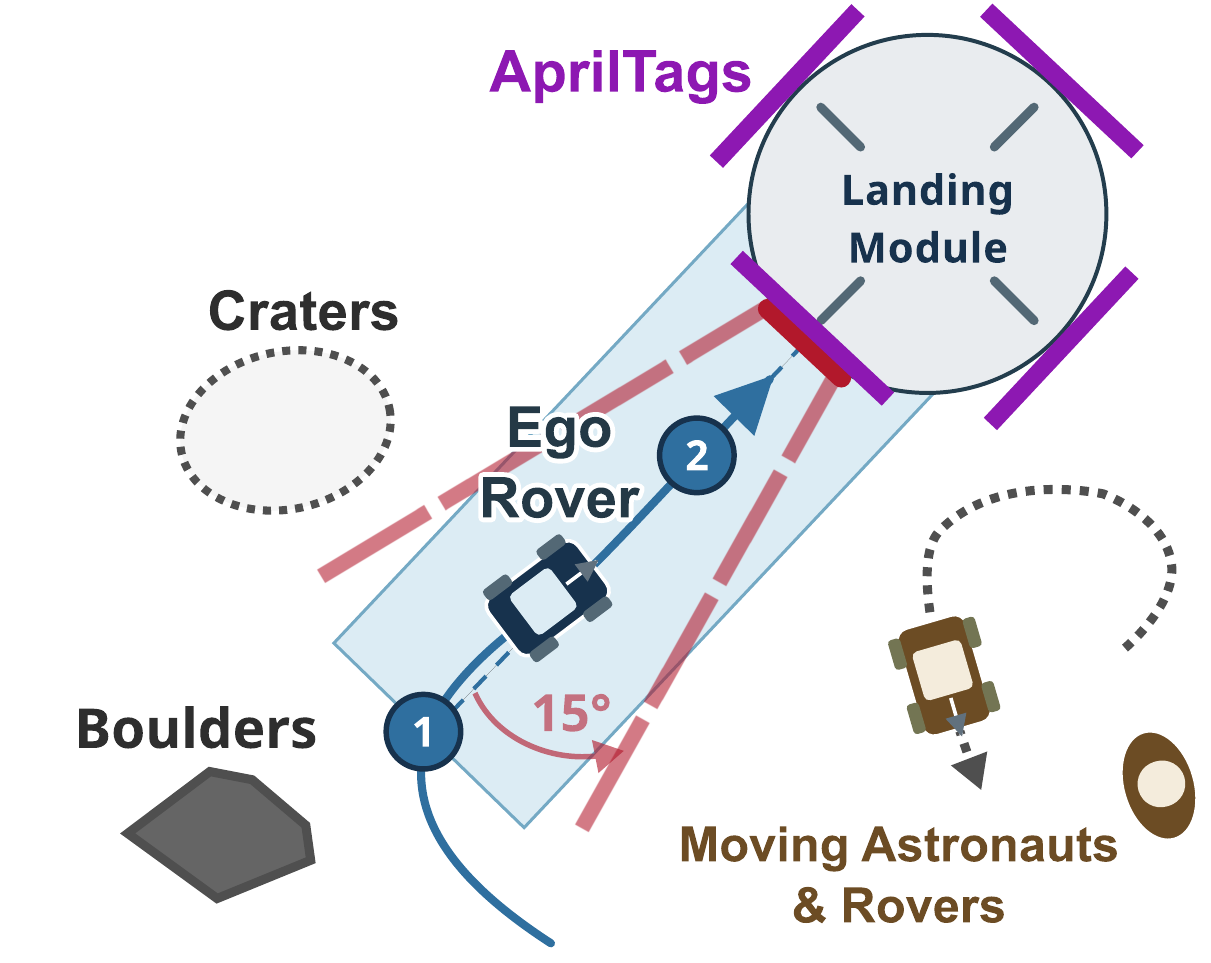}
        \caption{Schematic of the docking task and hazard landscape.}\label{fig:schematic:a}
    \end{subfigure}%
    ~ 
    \begin{subfigure}[h!]{0.50\textwidth}
        \centering
        \includegraphics[width=0.95\textwidth]{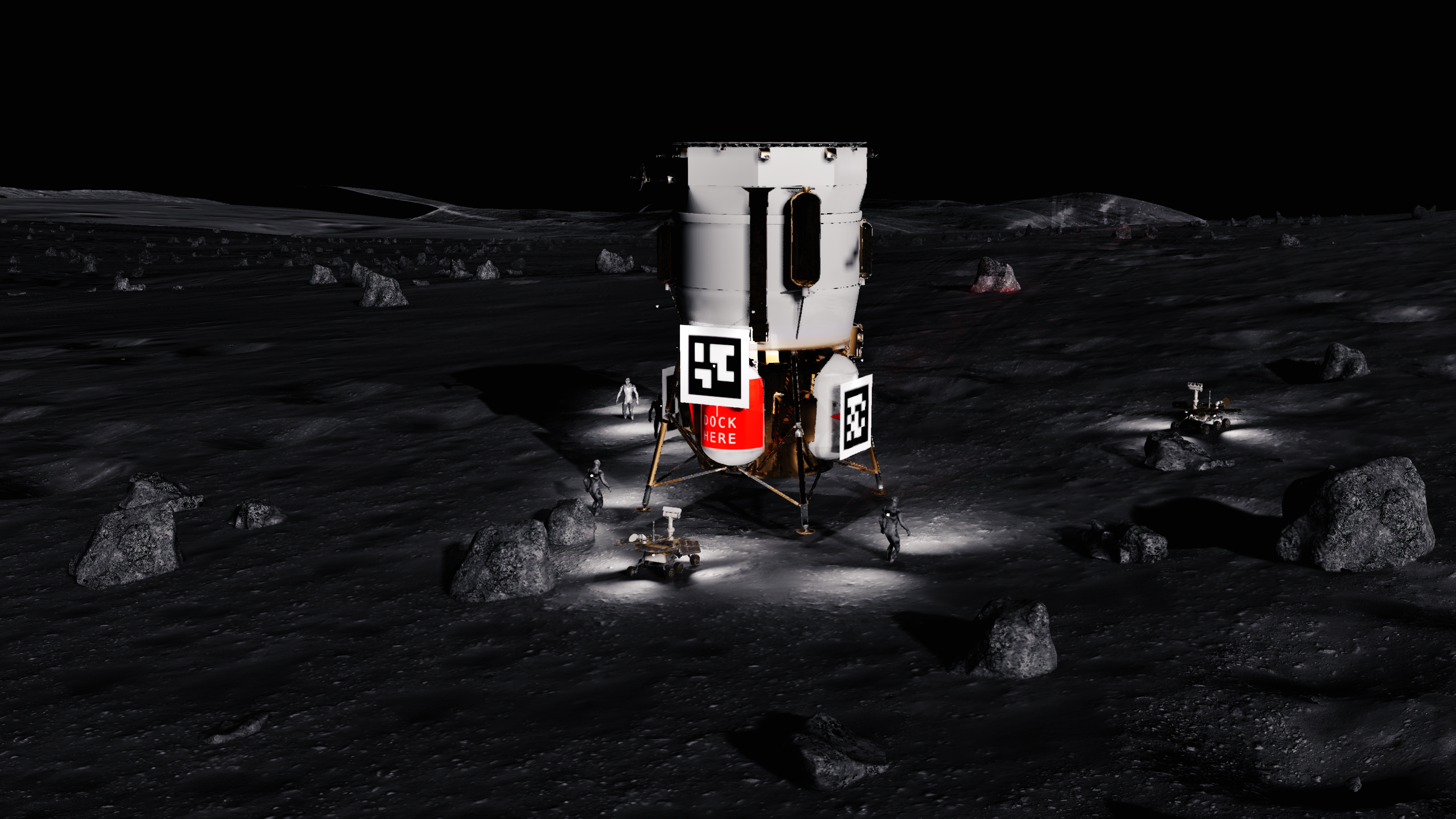}
        \caption{The LuPNT~\cite{vila_lupnt_2025} lunar surface simulation environment.}\label{fig:schematic:b}
    \end{subfigure}
    \caption{\textbf{Overview of the docking task.}
    The ego rover must enter the final-approach corridor (blue strip) then align with the landing module's docking port (red panel).
    Terrain, sun elevation and hazard landscape are randomized per scenario. 
    Docking is successful when the rover makes contact within 350 s and with $<15^\circ$ bearing error relative to the docking-port axis.
    }\label{fig:schematic}
\end{figure*}

Planetary rover autonomy has traditionally employed modular perception, mapping, and planning pipelines~\cite{biesiadecki_mars_2006,carsten_global_2007,toupet_enhanced_2026}. This provides interpretable intermediate representations and explicit safety constraints but can propagate perception errors into downstream planning~\cite{howard_vision-based_2001,li_transcendental_2023,hu_review_2025}.
End-to-end policies instead learn a direct mapping from perceptual inputs to control actions.
Because perception and control are optimized jointly, these policies need not produce explicit intermediate outputs, thus reducing failures at component interfaces.
However, deployment-time lighting and terrain appearance can differ from learned visual features and degrade performance~\cite{hansen_pre-training_2023,garcia_robust_2023}.
Mitigating this typically requires training on sufficiently large and diverse datasets~\cite{garcia_robust_2023,collaboration_open_2025,walke_bridgedata_2023}.

Vision-language-action (VLA) policies offer an alternative
by finetuning vision-language models (VLMs) trained on internet-scale image-text corpora for closed-loop control.
Such pretraining provides robust generalization over unseen scenarios~\cite{kim_openvla_2024,black__0_2024,sun_drop-then-recovery_2026}. In low-data regimes, improved data efficiency is associated with vision-language pretraining~\cite{li_vision-language_2024,wang_imitation_2026}.
To date, the application of end-to-end VLA policies to lunar surface navigation has yet to be investigated.
In this paper, we consider the scenario of autonomous docking between a lunar transport rover and a landing module.
We study whether VLA policies provide a robust alternative to modular pipelines in this context.

Despite vision-language pretraining, policy learning still requires high-quality demonstrations of safe navigation.
This is challenging first due to the infeasibility of large-scale data collection on the lunar surface; second, while physical testbeds and digital environments exist to simulate lunar conditions, collecting human-driver trajectories is costly and difficult to scale. 
We therefore adopt the Learn-by-Cheating (LBC) paradigm~\cite{chen_learning_2019} using privileged simulator information to generate expert demonstrations.
Our safety-aware expert encourages clearance from terrain and dynamic hazards while avoiding poorly illuminated regions.
Using this scalable data generation pipeline, we train a VLA policy that autonomously docks a lunar rover with a landing module, using visual observations from a single monocular camera, without prior maps of the environment.

\begin{journal}
Onboard computation is a constrained resource for lunar rovers. Reducing model inference frequency is attractive to limit computational demand and power consumption.
Our VLA policy adopts action chunking~\cite{zhao_learning_2023} to predict multiple navigation waypoints into the future.
We further introduce a physically grounded adaptive-horizon execution strategy that estimates how far the rover can safely travel in open loop.

Accurate and fine-grained trajectory corrections are necessary during final docking alignment.
To improve the policy's docking precision, we introduce a new technique, \textit{geometric temporal ensembling}, which aggregates multiple trajectory candidates within a single inference and across successive inferences.
\end{journal}

We construct a modular navigation baseline and compare its performance with that of the VLA policy, using the same underlying planner used for policy training but operating on online-estimated hazard information without privileged ground truth.
We evaluate both systems in simulation under nominal conditions and under sensor disturbances unseen during training.

Our contributions are summarized as follows: 
\begin{itemize}
    \item We design a VLA navigation policy for autonomous lunar rover docking.
    \begin{conference}
    \item We introduce a scalable training framework using safety-aware demonstrations generated through privileged expert planning.
    \end{conference}
    \begin{journal}
    \item We introduce an adaptive-horizon execution strategy to reduce inference frequency while maintaining closed-loop safety.
    \item We introduce geometric temporal ensembling to improve the precision of rover docking through inference-time trajectory ensembling.
    \end{journal}
    \item We benchmark the VLA policy in closed loop against a modular autonomy baseline under nominal conditions and lunar-motivated sensor disturbances.
\end{itemize}

\section{Related Work}

\subsection{Vision-Language-Action Policies for Navigation}

Vision-language-action models have been applied to terrestrial robot navigation in unstructured environments~\cite{zhang_uni-navid_2025,cheng_navila_2025,hirose_omnivla_2025}.
Terrestrial navigation policies benefit from abundant and heterogeneous data sources.
Non-robot data can be crowdsourced from egocentric cameras mounted on pedestrians and bicyclists, or repurposed from YouTube tour videos~\cite{windecker_navitrace_2025,liu_citywalker_2025,hirose_lelan_2024}. %
Multi-robot datasets can be aggregated~\cite{collaboration_open_2025,hirose_omnivla_2025} for training robust policies across different embodiments.
No comparably large and diverse body of training data is currently available for lunar surface navigation.
Policies trained exclusively on terrestrial imagery may not generalize to the lunar environment.

\subsection{Learning Navigation from Privileged Experts}

Imitation learning trains end-to-end policies using expert demonstrations, but collecting task- and environment-specific human demonstrations can be costly~\cite{chen_empirical_2022,zhang_end--end_2021}.
To reduce dependence on human demonstrations, LBC~\cite{chen_learning_2019} uses privileged simulator information to construct automated experts that generate training data at scale~\cite{zhang_end--end_2021,ozaibi_end--end_2024,adang_singer_2025}.
Importantly, privileged information need not be recoverable from deployment-time sensor observations.
In SPOC~\cite{ehsani_spoc_2024}, a learner policy trained from a privileged shortest-path planner exhibits robust closed-loop behaviors and outperforms one trained from an expert restricted to observable information.
However, shortest-path demonstrations do not account for deployment-time perceptual uncertainties and imperfect trajectory following.
Our expert generates demonstrations that encourage hazard clearance and avoid perceptually uncertain terrain, providing margins against both trajectory-following errors and perceptual uncertainty.

\begin{journal}
\subsection{Action Chunking and Adaptive-Horizon Execution}

Action chunking predicts a sequence of future actions per inference.
This reduces the effective decision frequency and can mitigate learner-induced compounding errors in imitation learning~\cite{zhao_learning_2023}.
To preserve closed-loop responsiveness, recent VLA methods adapt the execution horizon using online reinforcement learning~\cite{zhao_dynamic_2026}, model uncertainty~\cite{liang_adaptive_2026}, attention structure~\cite{wang_vla_2026}, action consistency~\cite{chen_dynamic_2026}, or observed execution drift~\cite{pan_vla-corrector_2026}.
While these signals estimate the reliability of a predicted sequence, they do not ground the horizon to the physical geometry of the environment.
Our VLA policy predicts curvature-conditioned execution horizons from visual observations and uses this prediction to determine the execution horizon.
\end{journal}

\subsection{Modular Planetary Rover Navigation}

Several works have sought to improve the robustness of individual components in modular navigation pipelines by replacing hand-engineered estimators with machine-learning models for localization and traversability assessment~\cite{tian_lo-slam_2025,liu_interplanetary_2026}.
In prior work~\cite{dai_full_2025}, we have developed and validated a lunar rover autonomy stack for hazard mapping and localization under lunar conditions.

\section{Methodology}

\subsection{Docking Task and Simulation Setup}

The lunar surface simulator LuPNT~\cite{vila_lupnt_2025}, built on Unreal Engine, is used to generate photorealistic imagery and simulate closed-loop vehicle dynamics~(Fig.~\ref{fig:schematic:b}).
Docking scenarios are 50~$\times$~50 m regions sampled from a 5 m/px resolution digital elevation model, each containing 6.3~$\pm$~4.1 (mean~$\pm$~SD) boulders (diameter: 2.4~$\pm$~0.8~m) and
5.0~$\pm$~3.6 craters (diameter: 4.0~m; depth: 0.8~$\pm$~0.1~m) on average.
The sun elevation is sampled from 20$^\circ$ to 90$^\circ$.
Moving hazards consist of astronauts and rovers (speed: 0.5~m/s), which are modeled to be cooperative and will yield when in close proximity with the rover.
The ego rover begins with the landing module in view at a distance of up to 50~m.
Four fiducial markers (AprilTag~\cite{wang_apriltag_2016}; width: 2~m) are attached to the sides of the landing module and artificially illuminated~(Fig.~\ref{fig:schematic}).
A 6-m-wide and 14-m-long hazard-free approach corridor is available on the docking port side, which is painted red~(Fig.~\ref{fig:schematic:b}). The ego rover is equipped with front-facing stereo cameras (height: 1.4~m above ground; baseline: 0.2 m;  field of view: 120$^\circ$; resolution: 1280~$\times$~720~px), with maximum velocities of $\pm$0.8~m/s.
The rover is placed at a randomized pose relative to the landing module, and must drive to contact the docking port within a relative bearing error of $15^\circ$~(Fig.~\ref{fig:schematic:a}) within 350~s of simulation time.
{%
Hazard contact is defined as collisions with boulders, astronauts, or other rovers, as well as crater incursions that exhaust the rover's 0.4~m ground clearance or cause a maximum possible wheel drop~\cite{toupet_enhanced_2026} greater than 0.4~m.}

\subsection{Training Data Synthesis}

An information-privileged expert planner generates safe docking trajectories and teaches recovery strategies from unsafe vehicle states.
Path optimization is performed using the Hybrid A* algorithm~\cite{dolgov_path_2010} with two-dimensional Reeds-Shepp curves~\cite{reeds_optimal_1990} as the rover's kinematics model.
To account for uncertainties in perceived hazards and imperfect trajectory following, the Voronoi safety field~\cite{meng_improved_2023} is adopted:
\begin{equation}
\begin{aligned}
\rho_V(x,y)
&= \left(\frac{\alpha}{\alpha+d_O(x,y)}\right)
   \left(\frac{d_V(x,y)}{d_O(x,y)+d_V(x,y)}\right) \\
&\quad
   \left(\max\left(0,\,
   1-\frac{d_O(x,y)}{d_O^{\max}}\right)\right)^2
\end{aligned}
\label{eq:voronoi}
\end{equation}
where $d_O(x,y)$ is the distance to the nearest hazard;
$d_V$ is the distance to the nearest edge of the generalized Voronoi diagram formed from terrain and moving hazards; $\alpha>0$ specifies the field's decay; and $d_O^{\max}>0$ limits the influence range of each hazard~\cite{meng_improved_2023}.
Intuitively, $\rho_V$ is minimized when the path lies on the medial axis between obstacles.

\begin{figure}
    \centering
    \includegraphics[width=\linewidth]{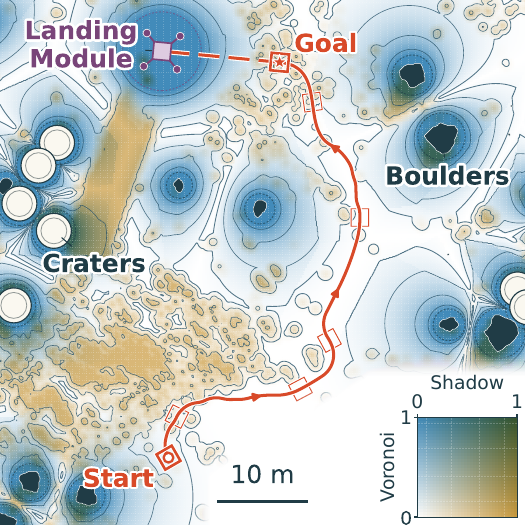}
    \caption{{\textbf{Representative cost map for expert-trajectory generation.}
    Voronoi safety field (blue) and shadow costs (orange) are visualized, with isocontours overlaid and darker regions denoting higher combined cost \(\rho(x,y)\).
    The expert planner generates a trajectory (solid red line) to a waypoint labeled ``Goal" at the entrance to the final-approach corridor. From the ``Goal" waypoint, the rover tracks the docking-port axis (dashed red line) to complete docking. }}
    \label{fig:cost_field}
\end{figure}

To discourage travel through poorly illuminated terrain,
a shadow cost $\rho_S(x,y)\in[0,1]$ is incorporated that scales with the local terrain luma.
The hazard and shadow costs are combined using the probabilistic OR operator as follows.
\begin{equation}\label{eq:shadow_voronoi}
\begin{aligned}
A(x,y) &= w_H {\rho}_V(x,y)\\
B(x,y) &= w_S \rho_S(x,y)\\
\rho(x,y)
&= A(x,y)+B(x,y)-A(x,y)B(x,y)
\end{aligned}
\end{equation}
where $w_H=1$ and $w_S=0.9$ are tunable weights of the Voronoi safety field and shadow cost, respectively.
This ensures the overall $\rho(x,y)$ is bounded within $[0,1]$ and is strictly increasing in either component for $\rho(x,y) < 1$.
A representative cost map of the combined $\rho(x,y)$ is visualized in Figure~\ref{fig:cost_field}.

Following ~\cite{meng_improved_2023}, the overall cost function for the Hybrid A* path planning is as follows.
\begin{equation}\label{eq:expert_search_cost}
f(x,y)
= w_g g(x,y)+w_h h(x,y)+w_v \rho(x,y)
\end{equation}
where $w_g$, $w_h$ and $w_v$ are tunable weights, and $g$ and $h$ are the motion and heuristic costs, respectively.

Before the final-approach corridor, the expert replans trajectories at 1 Hz, which are tracked by a PID controller~(Fig.~\ref{fig:cost_field}).
Upon arrival at the corridor, the controller tracks the approach axis perpendicular to the docking surface~(Fig.~\ref{fig:cost_field}).
To improve robustness of the learned policy, sun elevation, hazard placements, and the paths taken by moving hazards are randomized.
To teach expert recovery from deployment-time mistakes induced by the learner policy~\cite{ross_reduction_2011}, near-failure rover states (e.g., prior to crashing into a boulder) are oversampled as initial conditions.

\begin{journal}
To reduce inference frequency while preserving responsiveness to hazards, the VLA policy is supervised to predict how far the rover can travel before replanning.
Each training sample is annotated with eight execution horizons corresponding to maximum path curvatures:
\begin{equation}\label{eq:curvatures}
    \kappa_i\in\{0.125,0.25,\dots,1.0\}\,\text{m}^{-1}.
\end{equation}
For each $\kappa_i$, a Dubins reachable tube determines the minimum travel distance at which any admissible motion can intersect with a hazard or unobserved region, given a maximum rover speed of 0.8~m/s.
\end{journal}

Rover docking requires repositioning for the final approach while maintaining clearance from hazards. During such maneuvers, previously observed hazards can move outside the current field of view.
While the policy could in principle retain this information by integrating camera observations over time, learning implicit spatiotemporal representations may be challenging~\cite{chabal_fom-nav_2025}.
Our policy is provided with a bird's-eye-view (BEV) map as an explicit representation of past observations and rover trajectory.
The map is constructed online by projecting successive camera frames onto an assumed ground plane using a homography transformation~\cite{mallot_inverse_1991}.
The projections and rover positions are registered with simulated odometry in a rolling map centered on the rover~(Fig.~\ref{fig:arch}b).
In a physical implementation, the position and attitude estimates may be obtained by fusing wheel odometry with IMU measurements.
During policy training and evaluation, the position increments and attitude estimates are perturbed with noise to simulate odometry drift and sensor disturbances.

Camera frames,
\begin{journal}
execution horizons,~%
\end{journal}%
BEV maps,
and rover trajectories are recorded as training samples.
To assess robustness to deployment conditions unseen during training, the VLA policy is trained on a sterile dataset sans visual augmentation.
In total, 192k samples~(${\sim\!53}$~hr) of trajectory-observation pairs are collected over 2787 procedurally generated scenarios.

\subsection{VLA Policy Architecture and Training}

\begin{figure*}[h!]
    \centering
    \begin{journal}
    \includegraphics[width=\linewidth]{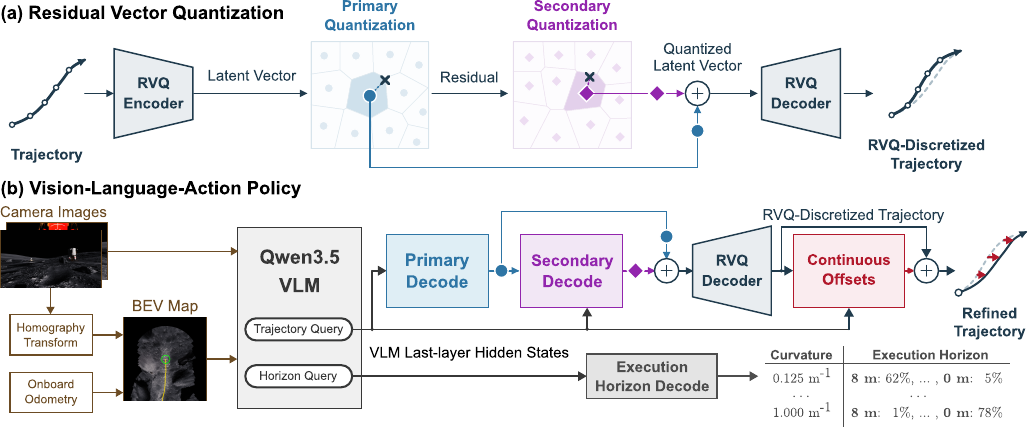}
    \end{journal}
    \caption{\textbf{Overview of the VLA policy and training.} a) Prior to policy training, an RVQ-VAE is trained on expert trajectories from the policy's training dataset, which then assigns each trajectory in the same dataset a primary and secondary code pair.
    b)
    The policy is trained to predict these primary and secondary codes from camera and BEV-map images. The quantized latent vector corresponding to predicted codes is decoded back to an RVQ-discretized trajectory, whose heading angles are refined by continuous offsets.
    \begin{journal}
    An execution-horizon head predicts a categorical distribution over execution horizons for each curvature bound. At deployment, the maximum curvature demanded for closed-loop trajectory tracking selects the corresponding execution horizon.
    \end{journal}}\label{fig:arch}
\end{figure*}

{Our VLA policy uses Qwen3.5-4B as the VLM backbone for its strong performance in visual understanding and spatial reasoning~\cite{qwen_team_qwen35_2026}.}
From a single monocular camera, the VLM backbone takes a rolling window of up to 60~s of image frames sampled at 0.33~Hz and a 640~$\times$~640~px BEV map as its conditioning inputs~(Fig.~\ref{fig:arch}).
{We downsample the camera frames to 512~$\times$~288~px to conserve memory and compute usage during training and inference.
Similar resolutions have been reported in prior work~\cite{nvidia_alpamayo-r1_2026,kim_openvla_2024}.}

{
Since the expert planner has access to information unavailable to the learner, similar learner observations can correspond to different underlying states and expert actions. This can increase multimodality in the conditional action distribution.
In our earlier experiments, continuous-valued waypoint predictions trained with an L1-regression objective~\cite{kim_fine-tuning_2025} perform poorly, which may reflect more pronounced multimodality in this dataset.
We therefore adopt the Vector-Quantized Behavior Transformer architecture~\cite{lee_behavior_2024} for action generation, which supports multimodal action distributions without the higher sampling latency of diffusion or flow matching methods.
}

The policy generates a trajectory of ten waypoints spaced 0.8~m apart, with each waypoint constructed from the preceding waypoint using a continuous-valued heading angle referenced to the ego orientation at the start of the trajectory.
Following~\cite{lee_behavior_2024}, we first train a variational autoencoder (VAE) with residual vector quantization (RVQ) on trajectories from the policy's training dataset.
{This discretizes the action space using primary and secondary code pairs.
The trained RVQ encoder then assigns each trajectory in the dataset corresponding primary and secondary codes.}
The primary and secondary codebooks each contain 24~codes, yielding $24\times24$ code combinations. A separate 25th primary code represents a 0.8~m reverse trajectory.

During inference, \begin{journal}trajectory and execution-horizon query tokens are\end{journal}
 appended to the VLM input. %
Following~\cite{lee_behavior_2024}, the primary-code prediction head (parameters: 19.8M) predicts primary-code probabilities from the trajectory query's hidden state; the secondary head (parameters: 3.8M) predicts secondary-code probabilities conditioned on both this state and a one-hot encoding of the sampled primary code.
Using the RVQ-VAE, the resulting code pair is decoded into a trajectory, which is refined by a continuous offset head (parameters: 19.7M) conditioned on the hidden state and the decoded trajectory.
\begin{conference}

\end{conference}
\begin{journal}
The execution-horizon query's hidden state is decoded by a separate head (parameters: 19.9M) that predicts the probabilities of discrete distance bins ($[0,1),\ldots,[7,8),[8,\infty)$~m) for each $\kappa_i$ in Eq.~\ref{eq:curvatures}.
\end{journal}

\begin{journal}
The policy's generated trajectory is tracked by a pure-pursuit controller. For a lookahead point at distance $L$ and bearing error $e$, the connecting arc requires curvature $\kappa = 2|\sin e|/L$. The controller retains the $\kappa_i$ nearest to the maximum required curvature since the last inference and uses the policy's predicted execution horizon associated with that $\kappa_i$.
Policy inference is triggered when the rover's accumulated travel exhausts the execution horizon. Empirically, we find that imposing a minimum 2~m horizon is sufficient for ensuring closed-loop responsiveness without introducing unnecessary inferences.
\end{journal}

During training, the primary%
\begin{journal}
,
\end{journal}
\begin{conference}

\end{conference}
secondary%
\begin{journal}
, and execution-horizon
\end{journal}
heads are supervised using focal loss~\cite{lin_focal_2018}.
The continuous offset head is supervised using mean squared error.
We balance these losses using homoscedastic uncertainty weighting~\cite{kendall_multi-task_2018}.
The overall loss function is as follows.
\begin{conference}

\end{conference}
\begin{journal}
\begin{equation}
\mathcal{L}
=
\sum_{\substack{
t\in\{\mathrm{primary},\,\mathrm{secondary},\\
\mathrm{execution\text{-}horizon},\,\mathrm{offset}\}
}}
\left(
c_t e^{-s_t}\mathcal{L}_t+\frac{1}{2}s_t
\right)
\end{equation}
\end{journal}

where $\mathcal{L}_t$ is the loss for task $t$ and $s_t$ is the log variance of the task-specific observation noise~\cite{kendall_multi-task_2018}; $c_t$ is set to $1$ and $\frac{1}{2}$ for classification and regression tasks, respectively.
The VLM's vision encoder and language backbone are finetuned using LoRA adapters~\cite{hu_lora_2021}~(rank: 142; parameters: 346M).
The adapters, query-token embeddings, and prediction heads are trained jointly for 3~epochs (AdamW; learning rate: $2\times10^{-5}$; weight decay: $10^{-5}$). %
A learning rate of $5\times10^{-3}$ is used to optimize the log variances.

\begin{journal}
\subsection{Geometric Temporal Ensembling}
Temporal ensembling~\cite{zhao_learning_2023} is a technique that aligns the actions by their timesteps and averages them across inferences, which improves the smoothness of the generated action chunks.
However, timestep alignment is not naturally suited to trajectory predictions, since corresponding prediction indices across successive inferences need not represent the same location relative to the rover.
Furthermore, during final docking, fine-grained corrective trajectories form primarily a unimodal distribution, and splitting modes may reduce docking precision.
We therefore propose geometric temporal ensembling (GTE), which aligns trajectories at fixed distances ahead of the rover's current position along each trajectory.
At each inference step, the top-$K$ trajectory candidates are averaged, weighted by their predicted probabilities.
Across successive inference steps, these aligned trajectories are further averaged and weighted by recency~\cite{zhao_learning_2023}.
The ensemble trajectory is then tracked by the PID controller.

The VLA policy is supervised to predict a zero execution-horizon target for the least-curvature class when the docking side is visible and the expert planner senses a stabilized approach. This makes the execution-horizon profile non-monotonic over curvature. At deployment, this pattern activates a final-docking mode, which applies GTE and queries the policy at an increased frequency.
\end{journal}

\subsection{Modular Autonomy Stack}

The modular autonomy baseline uses the same expert Hybrid A* planner for training data synthesis, with the hazards reconstructed using onboard observations.
A backup recovery planner is used when Hybrid A* cannot find a path towards the docking port. This planner ranks a set of forward and reverse arcs by hazard clearance and distance to the goal. If hazard clearance is less than 1.5~m, the backup planner ranks candidates by clearance alone.

We follow the methods described in~\cite{dai_full_2025} for front-end simultaneous localization and mapping (SLAM). Visual odometry is implemented using SuperPoint~\cite{detone_superpoint_2018} and LightGlue~\cite{lindenberger_lightglue_2023} for feature detection and matching, respectively. Stereo features are triangulated into 3D landmarks, whose matches in the next frame solve the relative pose transform via Perspective-n-Point~(PnP).
RAFT-Stereo~\cite{lipson_raft-stereo_2021} is used for dense stereo reconstruction due to its strong zero-shot performance reported on planetary rover datasets~\cite{liu_interplanetary_2026}.
The estimated stereo depths are used to generate dense 3D point clouds.
A U-Net++ model trained in LuPNT is selected for its strong image segmentation performance among the models evaluated in~\cite{dai_full_2025}.
{Semantic segmentation associates image regions with valuable semantic labels for navigation (e.g., boulders, landing module, and rovers~\cite{dai_full_2025}), and has been proposed for identifying regolith types suitable for traversal~\cite{toupet_enhanced_2026}.}

The traversability of grid cells is assessed using both geometric and semantic criteria.
Inspired by the Perseverance rover's ENav algorithm~\cite{toupet_enhanced_2026},
geometric traversability is assessed based on the cell surface tilt, elevation variation, and deviation from a fitted local ground plane.
Based on tunable thresholds, each cell is classified as clear, uncertain, or impassable.
Uncertain cells have elevated traversal costs.
{Semantic labels are combined with elevation criteria to reclassify uncertain cells as impassable.}
All parameter settings are tuned on $N=20$ scenario seeds outside the evaluation set until a 100\% docking success rate is achieved.

Docking-port localization is performed in two stages. Prior to detection of an AprilTag fiducial, the semantic point clouds labeled as the landing module are used to fit a bearing sector for coarse navigation toward the landing module. Once AprilTags are detected, the full pose of the docking port becomes available for the final docking approach.

\section{Evaluation}
\begin{figure}
    \centering
    \includegraphics[width=1.0\linewidth]{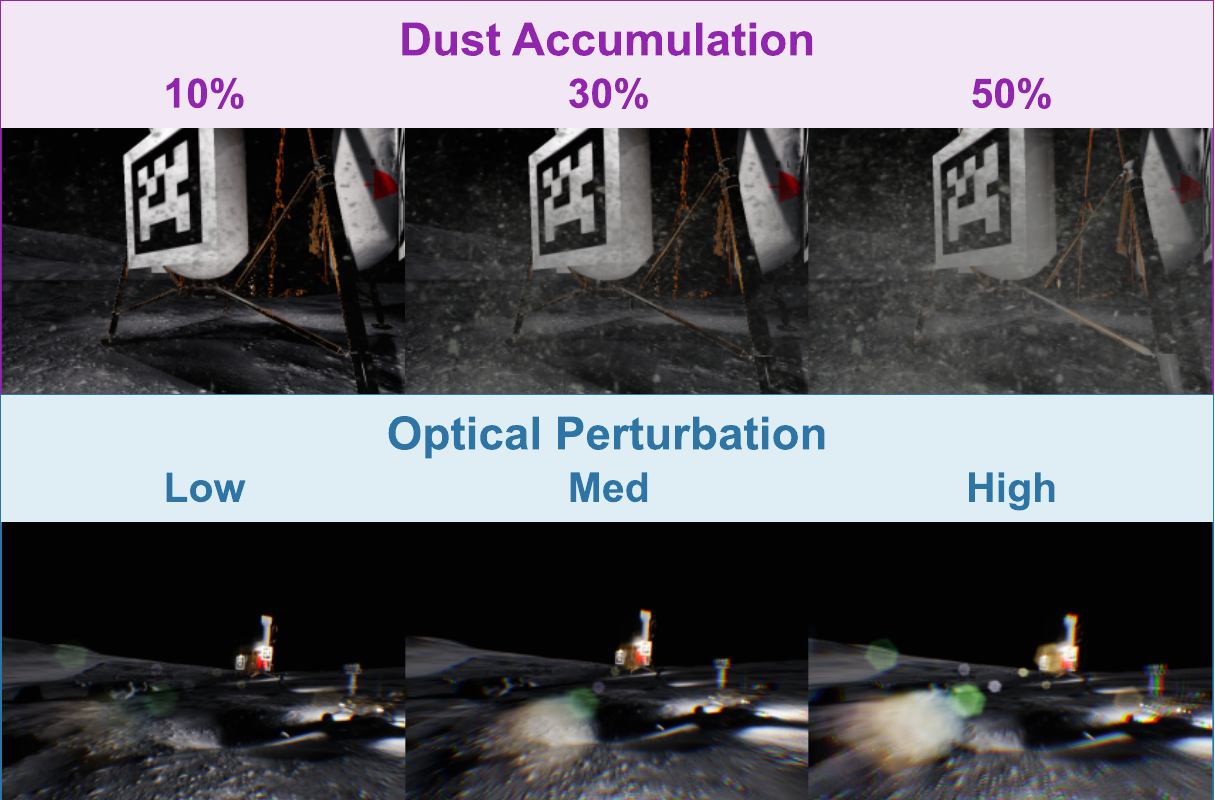}
    \caption{\textbf{Conditions evaluated under sensor disturbance.}}
    \label{fig:ablation_pics}
\end{figure}

Unless otherwise specified, evaluations are conducted in $N=100$ procedurally generated scenarios using seeds unseen during training-data synthesis, with initial rover-to-landing-module separations of 30--50~m. The VLA policy uses a trajectory sampling temperature of 0.26 and adaptive-horizon execution unless otherwise noted.

\subsection{Nominal and Sensor Disturbance Evaluations}

The VLA policy and modular autonomy stack are evaluated under the following nominal and sensor-disturbance conditions~(Fig.~\ref{fig:ablation_pics}). Since camera observations provide essential navigation information for both systems, visual perturbations are a primary source of perceptual artifacts expected at deployment~\cite{rodriguez-martinez_spice-hl3_2026,mantovani_defeating_2025,gaier_effects_2005}.
\begin{enumerate}
\item \textbf{Nominal condition:} The rover navigates to and docks with the landing module while avoiding hazards.
\item \textbf{Dust conditions:} Dust masks are overlaid at increasing opacity levels of 10\%, 30\% and 50\%, averaged over the image. Dust patterns are randomized per scenario.
\item \textbf{Optical conditions:} Bloom, lens flare, and chromatic aberration are simultaneously applied at Low, Med, and High effect strengths. For Low, the Unreal Engine settings are 2 for bloom intensity, 5 for lens flare intensity, and 2 for scene fringe intensity; for Med, they are 4, 10, and 4; and for High, they are 6, 15, and 6, respectively. Parameter settings are unitless.
\end{enumerate}

Across the Dust and Optical conditions, the performance of the VLA policy is most competitive relative to the modular baseline under the strongest visual perturbations~(Fig.~\ref{fig:dust_success_rate}).
These perturbed conditions are out-of-distribution (OOD) with respect to the training data, which comprise only unperturbed scenarios sans visual augmentation.

\begin{figure}
    \centering
    \begin{conference}
    \includegraphics[width=\linewidth]{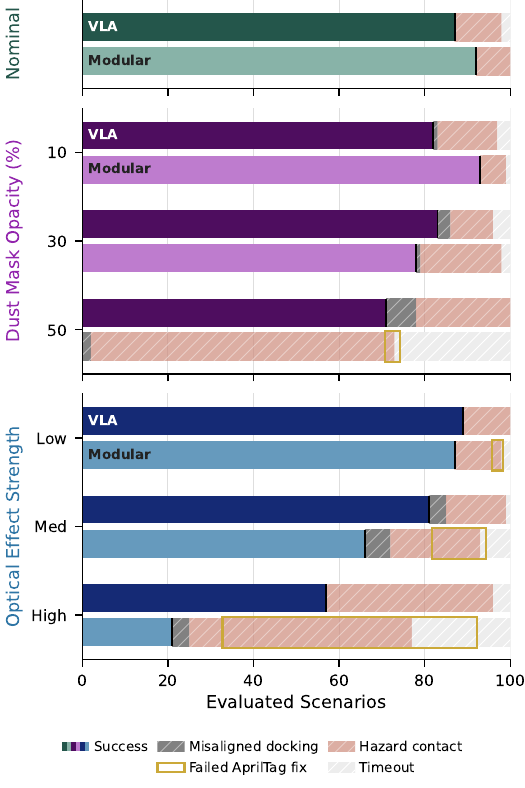}    
    \end{conference}
    \includegraphics[width=\linewidth]{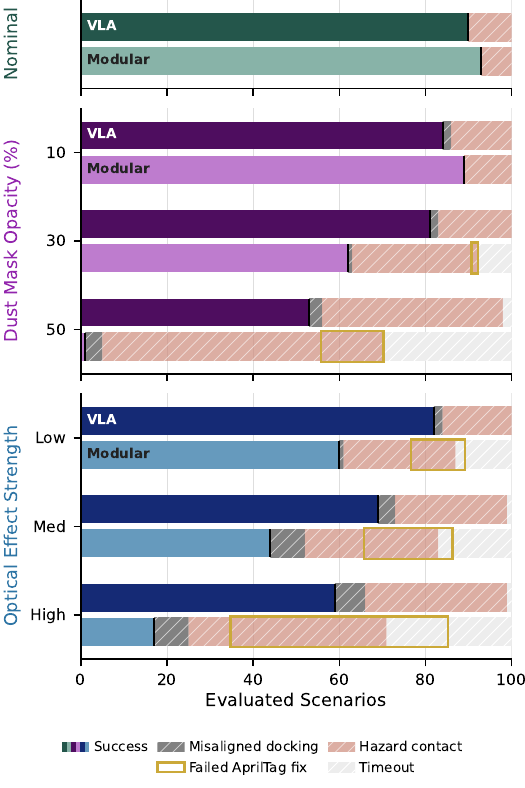}
    \caption{\textbf{Docking success rates under nominal and sensor disturbance conditions.} The VLA policy demonstrates reduced sensitivity to sensor disturbances relative to the modular baseline.}
    \label{fig:dust_success_rate}
\end{figure}

\begin{figure*}[b]
    \centering
    \includegraphics[width=\linewidth]{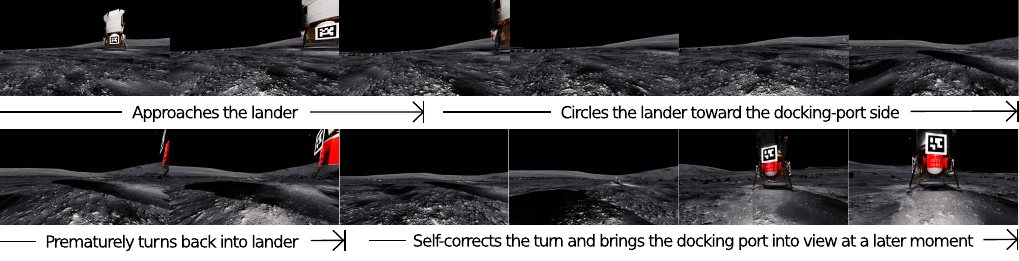}
    \addtocounter{figure}{1}
    \caption{\textbf{Emergent closed-loop behaviors not present in the expert demonstrations.}}
    \label{fig:emergent}
    \addtocounter{figure}{-2}
\end{figure*}
\begin{figure}[h]
    \centering
    \includegraphics[width=\linewidth]{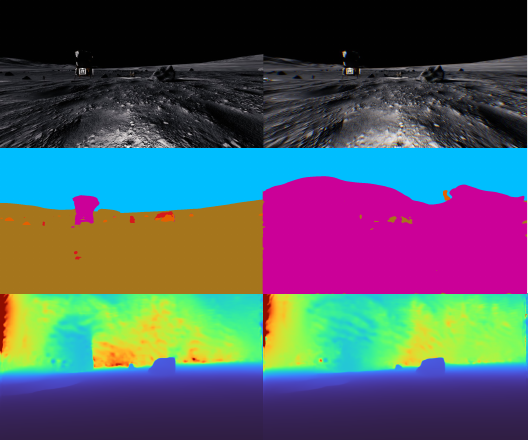}
    \caption{\textbf{Camera observations and derived perception estimates for the Nominal (left) and Low optical-effect-strength (right) conditions.} Degradations in stereo depth estimates (bottom) and semantic segmentation (middle) are observed.}
    \label{fig:slam_error_pics}
    \addtocounter{figure}{1}
\end{figure}

Hazard contact is a dominant failure mode for the VLA policy, accounting for 87\% (158/182) of unsuccessful scenarios.
Hazard contact becomes more frequent at the strongest perturbation conditions, accounting for 42 of 100 scenarios at 50\% dust opacity and 33 of 100 scenarios at High optical effect strength.
Training the policy to predict a STOP action under high perceptual uncertainty may help reduce hazard contact.
Simulation timeout accounts for 2\% (4/182) of all unsuccessful scenarios.
When the policy fails to dock within the allocated 350~s of simulation time,
repeated attempts for alignment near the docking port are observed.
Failure attribution can be difficult for VLA policies as their internal representations are not directly interpretable.
Future work should incorporate embodied reasoning~\cite{zawalski_robotic_2025} that verbalizes perceived hazards and high-level planning to aid failure interpretation and to further improve performance.

For the modular baseline, visual perturbations are associated with errors in docking-port localization and hazard mapping.
Docking requires AprilTag-based localization of the docking port, which fails in 50 of 83 unsuccessful scenarios at High optical effect strength.
Due to persistent odometry errors, 8\% of scenarios end with docking bearing errors above the $15^\circ$ requirement in the Med and High optical effect conditions.
These failures often coincide with spuriously mapped hazards, which stall the rover and prevent progress toward the docking port.
PnP odometry errors and degraded stereo depth can compromise hazard mapping during geometric traversability assessment~(Fig.~\ref{fig:slam_error_pics}).
Substantial semantic misclassifications are observed at perturbations as low as 10\% dust opacity, as well as at Low optical effect strength~(Fig.~\ref{fig:slam_error_pics}).
Hazard contact occurs in 65 of 100 scenarios at 50\% dust opacity and 46 of 100 at High optical effect strength.

The modular baseline's sensitivity to perturbations is consistent with both its perception training and reliance on geometric hazard reconstruction. U-Net++ is trained on sterile imagery in our simulator, which may explain its sensitivity to visual perturbations. Although RAFT-Stereo is trained with substantial photometric data augmentation~\cite{lipson_raft-stereo_2021}, the geometric traversability criteria remain highly sensitive to measurement errors. Direct trajectory prediction avoids propagating such errors through hazard mapping into planning. Future investigation is warranted on the effects of sensor disturbance on lunar visual SLAM.

The VLA policy's visual pretraining may reduce its sensitivity to perturbed observations~\cite{oquab_dinov2_2024}.
Language backbones are known to contribute additional robustness under visual and physical perturbations~\cite{sun_drop-then-recovery_2026}.
The individual contributions of pretraining and architecture are not isolated in this study.
Nevertheless, the policy's elevated hazard contact rates and incomplete docking approaches under perturbation indicate continued sensitivity to degraded observations.
This experiment evaluates the VLA policy under OOD conditions after training on a sterile dataset.
While visual augmentation is not applied in this study, future work may explore techniques such as conditional video generation~\cite{hui_seeing_2026} for training lunar policies on more diverse imagery.

\subsection{Emergent Behaviors}
Imitation of an information-privileged expert can produce closed-loop behaviors beyond those explicitly demonstrated by the expert~\cite{ehsani_spoc_2024}. Our VLA policy exhibits strategies unseen during training, including self-corrective recovery and peeking around occlusions. Consistent with~\cite{ehsani_spoc_2024}, this may be achieved by the composition of requisite subskills demonstrated in the expert data. In Figure~\ref{fig:emergent}, states reached after the premature turn can resemble the unsafe initial conditions that are sampled during training. These trajectories provide corrective actions, and the nominal demonstrations provide the continued circumnavigation and approach. Conditioned on past observations, the policy connects these behaviors through learner-generated histories containing its own preceding mistakes.
The temporally composed recovery sequence is emergent and absent from training.

\begin{journal}
\subsection{Effects of Adaptive-Horizon Execution}

We evaluate the effect of the execution horizon on docking performance. Fixed-horizon baselines at 2, 3, 4, 6, or 8~m are compared against our adaptive-horizon strategy.
Figure~\ref{fig:adaptive_hor} reports the docking-success and hazard-contact rates along with the mean inference calls per successful scenario.
Inferences during GTE activation are excluded from the analysis as the policy is queried at a higher fixed frequency during GTE.

Under fixed horizons, docking success decreases from 90\% at 2~m to 72\% at 8~m, while hazard contact increases from 9\% to 21\%, consistent with the benefits of frequent closed-loop feedback.
Although retaining predictions conditioned on earlier, less distribution-shifted observations can mitigate learner-induced error compounding~\cite{lazzati_why_2026},
docking success rate is observed to decrease with longer horizons over the evaluated conditions~(Fig.~\ref{fig:adaptive_hor}).
As each waypoint is constructed from the preceding ones, longer execution horizons increase exposure to errors propagated through successive waypoint predictions.
In the regime where accumulated action errors outweigh the benefit of conditioning on earlier observations, shorter execution horizons help to improve performance~\cite{zhao_dynamic_2026}.

The adaptive-horizon condition achieves 90\% docking success and 10\% hazard contact, compared with 90\% and 9\%, respectively, under the fixed 2-m horizon, while requiring 51.3\% fewer policy inferences per successful scenario (8.36 versus 17.16).
Relative to the fixed 6-m horizon, adaptive horizon increases docking success from 76\% to 90\% and reduces hazard contact from 15\% to 10\% at a similar inference cost (8.36 versus 7.96 calls).
Taken together, the performance of adaptive horizon suggests a potential in reducing deployment-time compute and power consumption.
Executing longer horizons where clearance allows can limit the consequences of erroneous predictions.
In regions with low hazard clearance, adaptive horizon allocates more inference calls to allow frequent closed-loop feedback and correction.

\begin{figure}
    \centering
    \includegraphics[width=\linewidth]{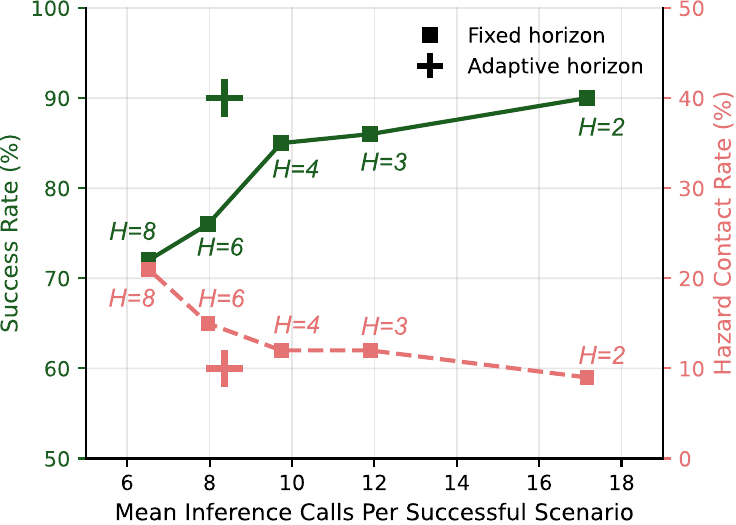}
    \caption{\textbf{Docking-success and hazard-contact rates over mean inference calls per successful scenario.} Squares denote fixed-horizon conditions (8, 6, 4, 3, and 2~m, from left to right); crosses denote the adaptive-horizon condition (mean distance traveled per inference call: 4.34~m). The adaptive-horizon condition achieves a similar docking-success rate (90\%) and hazard-contact rate (10\% versus 9\%) compared with the fixed 2-m horizon while requiring 51\% fewer inference calls per successful scenario.}
    \label{fig:adaptive_hor}
\end{figure}

\subsection{Effects of BEV Maps on Policy Performance}

The performance of the VLA policy is evaluated in open loop with and without BEV maps using $N=3203$ held-out validation samples.
A single model is jointly optimized under both input conditions by alternating
between including and omitting the BEV map on successive training
samples.
Execution horizons and trajectories are decoded greedily, with autoregressive trajectory decoding.
For each sample~\(i\) and curvature index \(c\in\{1,\ldots,C\}\), where \(C=8\), the ground-truth execution horizon $h_{i,c}$ is discretized as \(b_{i,c}=\min(\lfloor h_{i,c}\rfloor,8)\), and the predicted bin is denoted as \(\hat b_{i,c}\). The curvature-specific bin accuracy is defined below.
\begin{equation}
    \mathrm{Acc}_{c}=\frac{100}{N}\sum_{i=1}^{N}\mathbf{1}\!\left[\hat b_{i,c}=b_{i,c}\right].
\end{equation}
Aggregating over curvatures, the bin and overestimate-bin mean absolute errors (MAEs) are as follows.
\begin{align}
\mathrm{MAE}_{\mathrm{bin}}
&=\frac{1}{NC}\sum_{i=1}^{N}\sum_{c=1}^{C}
\left|\hat b_{i,c}-b_{i,c}\right| \\
\mathrm{MAE}_{\mathrm{over}}
&=\frac{1}{NC}\sum_{i=1}^{N}\sum_{c=1}^{C}
\max\!\left(\hat b_{i,c}-b_{i,c},0\right)
\end{align}
To characterize the upper tail of the overestimate-bin error distribution, the conditional value-at-risk (CVaR) at the 95\% level is computed as the mean overestimate-bin error in the worst 5\% of frame–curvature predictions as follows:
\begin{equation}
\mathrm{CVaR_{95}}\!
=\!\min_{\eta\geq 0}\!
\left[
\eta\!+\!\frac{1}{0.05NC}\!
\sum_{i=1}^{N}\!\sum_{c=1}^{C}
\max\!\left(\hat b_{i,c}\!-\!b_{i,c}\!-\!\eta,0\right)
\right]\!
\end{equation}
where \(\eta\) is the threshold selected at the 95th percentile of the overestimate-bin errors.

Let \(\mathcal{V}\) be the expert trajectory waypoints, and \(\theta_{i,j}\) and \(\hat\theta_{i,j}\) be the expert and predicted headings in degrees, respectively. Waypoint heading MAE is defined below.
\begin{equation}
\mathrm{MAE}_{\mathrm{heading}}
=\frac{1}{|\mathcal{V}|}
\sum_{(i,j)\in\mathcal{V}}
\left|\hat\theta_{i,j}-\theta_{i,j}\right|.
\end{equation}
To evaluate the open-loop hazard contact rates, the Expected Hazard Contact Rate (EHCR) is computed as follows.
\begin{equation}
    \mathrm{EHCR}=\frac{100}{N}\sum_{i=1}^{N}\sum_{k}p_{i,k}o_{i,k}
\end{equation}
where \(p_{i,k}\) is the policy's predicted probability of candidate trajectory \(k\), and \(o_{i,k}\in\{0,1\}\) indicates whether the predicted trajectory contacts a hazard.
Since sampling may be applied during trajectory decoding,
EHCR is chosen over hazard contact rates to better capture the likelihoods assigned to unsafe trajectories. EHCR is evaluated separately using the full predicted trajectories and by truncating each trajectory using the predicted execution horizons.

\begin{table}
\renewcommand{\arraystretch}{1.15}
\caption{\textbf{Execution-horizon and trajectory prediction metrics with and without BEV maps on open-loop evaluations ($\mathbf{N=3203}$).} BEV maps reduce hazard contact rates and waypoint heading MAE, and increase execution-horizon prediction accuracy, with larger improvements at higher curvatures. %
}
\label{tab:bev_ablation}
\centering
\footnotesize

\begin{tabular}{|l|c|c|}
\hline
&
\bfseries Without BEV &
\bfseries With BEV \\
\hline\hline

\multicolumn{3}{|l|}{\bfseries Execution horizon} \\
\hline

\multicolumn{2}{|l|}{
Bin accuracy (\%) by curvature ($\mathrm{m}^{-1}$) $\uparrow$} & \\
\quad 0.125 & 83.39 & \textbf{84.92} \\
\quad 0.250 & 79.46 & \textbf{82.24} \\
\quad 0.375 & 75.15 & \textbf{80.02} \\
\quad 0.500 & 73.84 & \textbf{80.61} \\
\quad 0.625 & 73.37 & \textbf{80.86} \\
\quad 0.750 & 73.37 & \textbf{81.11} \\
\quad 0.875 & 72.87 & \textbf{79.93} \\
\quad 1.000 & 72.62 & \textbf{79.55} \\
\quad Overall & 75.51 & \textbf{81.15} \\

Bin MAE (bins) $\downarrow$
    & 0.43 & \textbf{0.33} \\
Overestimate-bin MAE (bins) $\downarrow$
    & 0.21 & \textbf{0.16} \\
Overestimate-bin $\mathrm{CVaR_{95}}$ (bins) $\downarrow$
    & 2.92 & \textbf{2.57} \\
\hline

\multicolumn{3}{|l|}{\bfseries Trajectory} \\
\hline
Waypoint heading MAE ($^\circ$) $\downarrow$
    & 15.19 & \textbf{13.99} \\
EHCR (\%) $\downarrow$ &  &  \\
\quad Execution-horizon truncated & 2.13 & \textbf{1.89} \\
\quad Full trajectory & 10.71 & \textbf{9.96} \\
\hline

\end{tabular}
\end{table}

Including BEV maps increases execution-horizon bin accuracy from 75.51\% to 81.15\%, while reducing the MAE from 0.43 to 0.33~bins and the overestimate MAE from 0.21 to 0.16~bins~(Tab.~\ref{tab:bev_ablation}). Improvements in bin accuracy are consistent across curvatures, with more pronounced improvements (6.93--7.74~\%) for curvatures above 0.500~m$^{-1}$. Execution horizons account for both hazards and unobserved regions. At higher curvature bounds, the reachable tube extends farther laterally, which makes horizon prediction more dependent on regions outside the current camera view. This aligns with the larger accuracy improvements at higher curvatures and the lower overall overestimate MAE.
BEV maps improve hazard avoidance performance, with the execution-horizon truncated EHCR decreasing from 2.13\% to 1.89\%. Controlling for the effects of horizon predictions, the full trajectory EHCR is also reduced from 10.71\% to 9.96\%.  Including BEV maps decreases the heading MAE from the expert's reference trajectories ($15.19^\circ$ to $13.99^\circ$).

\subsection{Effects of GTE on Docking Accuracy}

$N=50$ paired scenarios with successful docking are selected for analysis. The following conditions are evaluated on the selected scenario seeds:
\begin{enumerate}
    \item \textbf{Baseline condition}: The docking trajectories are executed to the full extent permitted by the predicted execution horizon.
    \item \textbf{``A'' condition}:  A constant execution horizon of 1~m is applied during final docking mode.
    \item \textbf{``B'' condition}: The ``A" condition is augmented with likelihood-weighted ensembling of the top-$K$ trajectory candidates ($K=5$) within each inference.
    \item \textbf{``C'' condition}: The ``A" condition is augmented by aligning and averaging trajectory candidates across inferences, weighted by recency.
    \item \textbf{GTE condition}: The within-inference ensembled trajectories are further aligned and averaged across inferences, weighted by recency.
\end{enumerate}%

The final docking bearing angle errors for the paired successful scenarios under all five conditions are reported in Table~\ref{tab:gte_ablation}.
A Wilcoxon signed-rank test is used to assess GTE against the Baseline condition.
GTE significantly reduces the mean absolute docking bearing error by 35\%~(7.24$^\circ$ to 4.73$^\circ$; $p=0.015$; Cohen's $d_z=0.35$) while limiting the standard deviation to 54\% of the Baseline condition~(6.04$^\circ$ to 3.27$^\circ$).
More frequent replanning is associated with improved~($+$1.26$^\circ$) docking accuracy.
Ensembling trajectories within each inference appears beneficial when combined with ensembling across inferences.

\begin{table}
\renewcommand{\arraystretch}{1.15}
\setlength{\tabcolsep}{4pt}
\caption{\textbf{Docking bearing errors (mean $\pm$ SD; $N=50$
per condition) across ablated components of GTE.} GTE significantly improves docking accuracy relative to Baseline. *$p=0.015$}
\label{tab:gte_ablation}
\centering
\footnotesize

\begin{tabular}{|l|c|c|c|c|}
\hline
& &
\multicolumn{2}{c|}{\bfseries Trajectory ensembling}
& \\
\cline{3-4}
\bfseries Condition &
\shortstack{\bfseries Frequent\\\bfseries inference} &
\shortstack{\bfseries Within\\\bfseries inference} &
\shortstack{\bfseries Across\\\bfseries inferences} &
\shortstack{\bfseries Docking bearing\\
            \bfseries error ($^\circ$) $\downarrow$} \\
\hline\hline

Baseline  &   &   &   & $7.24 \pm 6.04$ \\  %
A         & X &   &   & $5.98 \pm 4.39$ \\ %
B         & X & X &   & $6.67 \pm 4.16$ \\ %
C         & X &   & X & $5.83 \pm 3.92$ \\ %
\hline
\textbf{GTE}
          & X & X & X & $\mathbf{4.73 \pm 3.27^*}$ \\  %
\hline

\end{tabular}
\end{table}

\subsection{Effects of Video-history Length}
The VLA policy is provided with a video of past camera observations, and limiting video-history length is desirable to lower inference latency and memory requirements.
To assess the effects of video-history length on closed-loop performance, video-history lengths of 5, 10, and 20 frames are evaluated. The docking-success and hazard-contact rates, and mean in-simulation docking completion time are presented in~Table~\ref{tab:video_history}.

Inconsistent trends in docking success and hazard contact rates are observed across the conditions~(Tab.~\ref{tab:video_history}).
The docking-success rates are 87\%, 84\%, and 90\% for the 5-, 10-, and 20-frame conditions, respectively, while the corresponding hazard-contact rates are 11\%, 14\%, and 10\%.
Neither metric shows a monotonic trend with increasing video-history length.
In contrast, the mean docking completion time decreases consistently from $114.7$~s with 5 frames to $93.5$~s with 10 frames and $81.4$~s with 20 frames. The variability in completion time also decreases with longer video histories, with the standard deviation declining from $62.9$~s to $40.7$~s and $25.3$~s, respectively.

Rover docking is a long-horizon task in which relevant visual features can remain outside the camera's field of view during repositioning around hazards and the landing module.
The observation history helps to preserve localization of the target and the preceding approach.
For instance, when the last frame containing the landing module falls outside the video-history window, we observe the policy often prioritizing a search pattern to reacquire the landing module that interrupts the preceding maneuver.
At the nominal speed of 0.8~m/s, a half-circumnavigation around the landing module takes approximately 32~s, during which the docking target is occluded from the rover's field of view~(Fig.~\ref{fig:emergent}).
At 0.33~Hz, the 5-, 10-, and 20-frame conditions provide approximately 15, 30, and 60~s of visual history, respectively.

\begin{table}
\renewcommand{\arraystretch}{1.15}
\setlength{\tabcolsep}{6pt}
\caption{\textbf{Closed-loop docking performance metrics evaluated in $N=100$ scenarios per condition.} Successful docking completion time is reported in mean~$\pm$~SD.}
\label{tab:video_history}
\centering
\footnotesize

\begin{tabular}{|c|c|c|c|}
\hline
& \multicolumn{2}{c|}{\bfseries Rate (\%)} & \\
\cline{2-3}
\shortstack{\bfseries Video-history\\
            \bfseries length (frames)} &
\shortstack{\bfseries Docking\\
            \bfseries success} &
\shortstack{\bfseries Hazard\\
            \bfseries contact} &
\shortstack{\bfseries Mean docking \\
            \bfseries completion time (s)} \\
\hline\hline

5  & 87 & 11 & 114.7~$\pm$~62.9 \\
10 & 84 & 14 & 93.5~$\pm$~40.7 \\
20 & 90 & 10 & 81.4~$\pm$~25.3 \\

\hline
\end{tabular}
\end{table}

\end{journal}

\subsection{Inference Latency}

\begin{table}
\renewcommand{\arraystretch}{1.15}
\setlength{\tabcolsep}{4pt}
\caption{\textbf{Policy inference latency (ms) data.} P95 denotes the 95th percentile.}
\label{tab:runtime}
\centering
\footnotesize

\begin{tabular}{|l|r|r|}
\hline
\bfseries Component & \bfseries Median & \bfseries P95 \\
\hline\hline
Input preprocessing & 41  & 56 \\
Qwen3.5-4B backbone        & 183 & 189 \\
Trajectory \begin{journal}+ execution-horizon~\end{journal}heads & 2   & 2 \\
\hline
Total               & 225 & 247 \\
\hline
\end{tabular}
\end{table}

All evaluations in this study are conducted on NVIDIA RTX~5090
and Intel Core Ultra~9~285K. The policy inference times are reported in Table~\ref{tab:runtime}. A single forward pass is required through the VLM backbone per inference~(183~ms). \begin{journal}Execution-horizon and trajectory sampling heads add negligible latencies~(2~ms).\end{journal}\begin{conference}\end{conference}
Near-real-time performance is achieved with a median inference latency of 225~ms.

A comparable platform for embedded hardware may use the NVIDIA Jetson AGX Thor T5000. To estimate the model latency on AGX Thor, the theoretical latency model from \cite{jiang2026fast} is adopted as follows.
\begin{equation}
T=\sum_{o\in O} \max\left(\frac{\mathrm{FLOPs}_o}{\mathrm{FLOP/s}},\frac{\mathrm{Bytes}_o}{\mathrm{MemBW}}\right)
\end{equation}
where \(o\) denotes a computational block in the model, \(\mathrm{FLOPs}_o\) and \(\mathrm{Bytes}_o\) denote the per-inference floating-point operations (FLOPs) and data transfer, respectively, and \(\mathrm{FLOP/s}\) and \(\mathrm{MemBW}\) denote the hardware's compute throughput and memory bandwidth.
The per-inference data transfer and FLOPs are measured using Nsight Compute~\cite{nvidia_corporation_nvidia_nodate} and PyTorch~\cite{paszke_pytorch_2019}, respectively.
The theoretical minimum latency $T$ is computed as 108~ms for RTX 5090 (compute-bound; 210 BF16 TeraFLOP/s; 1,792~GB/s; 575~W TDP) and 340~ms for AGX Thor (memory-bound; 250 BF16 TeraFLOP/s; 273~GB/s; 130~W TDP).
Future work can explore model quantization and distillation to further lower latency.

\section{Conclusion}

The aim of this study is to characterize the performance of end-to-end learned VLA policies for lunar surface navigation.
The VLA policy achieves higher docking success in five of the six evaluated sensor-disturbance conditions, with the largest advantages under the strongest perturbations.
The policy exhibits emergent behaviors absent in the training data, enabling robust self-corrective recovery of its deployment-time mistakes. %
\begin{journal}
To limit compute consumption, we propose an adaptive-horizon execution strategy that predicts how far a rover can safely travel in open loop before replanning.
This strategy achieves comparable docking-success rates while limiting inference frequency relative to a fixed-horizon baseline.~%
\end{journal}%
While this study considers the task of autonomous rover docking, rich and diverse task specifications can be described in natural language.
Future work will investigate multi-task versatility by further exploiting the broad language pretraining of VLA policies.

\acknowledgements %
This manuscript benefited from the use of AI assistants (GPT~5.6 and GPT~5.5) for manuscript revising and coding assistance. All final content was reviewed and edited by the authors. The authors thank Blue Origin for funding this project. The authors thank Adam Dai, Guillem Casadesus Vila, and Asta Wu for their valuable feedback and discussion.

\printbibliography[heading=bibintoc,title={References}]

\end{document}